\documentclass{article} % For LaTeX2e
\usepackage{nips14submit_e,times}
\usepackage{amsfonts}
\usepackage{amssymb}
\usepackage{amsmath}
\usepackage{bm}
\usepackage{natbib}
\usepackage{subfigure}
\usepackage{graphicx}
\usepackage{algorithm}
\usepackage{algorithmic}
\usepackage{threeparttable}
\usepackage{booktabs}
\usepackage{url}
\usepackage{color}

\nipsfinalcopy 

\title{Beta Process Non-negative Matrix Factorization \\with Stochastic Structured Mean-Field Variational Inference}
\author{Dawen Liang\\ Department of Electrical Engineering\\ Columbia University \\ \tt{dliang@ee.columbia.edu}
\And
Matthew D. Hoffman\\ Adobe Research \\ Adobe Systems Incorporated\\ \tt{mathoffm@adobe.com}}

\begin{document}

\maketitle

\begin{abstract}
Beta process is the standard nonparametric Bayesian prior for latent factor model. In this paper, we derive a structured mean-field variational inference algorithm for a beta process non-negative matrix factorization (NMF) model with Poisson likelihood. Unlike the linear Gaussian model, which is well-studied in the nonparametric Bayesian literature, NMF model with beta process prior does not enjoy the conjugacy.  We leverage the recently developed stochastic structured mean-field variational inference to relax the conjugacy constraint and restore the dependencies among the latent variables in the approximating variational distribution. Preliminary results on both synthetic and real examples demonstrate that the proposed inference algorithm can reasonably recover the hidden structure of the data.
\end{abstract}

\section{Introduction}

Non-negative matrix factorization (NMF) model, which approximately decomposes a non-negative matrix into the product of two non-negative matrices (usually referred as the latent component and the activation), is widely used in many application domains, such as music signal analysis \citep{smaragdis2003non} and recommender systems \citep{gopalan2013scalable}. One hyperparameter in the NMF model is the number of latent components, which is usually set via model selection (e.g. cross validation). Nonparametric Bayesian latent factor models, on the other hand, offer an alternative solution by putting an infinite-dimensional prior on the latent component and activation matrices, and allow the data to ``speak for itself'' via posterior inference. 

Most of the literature on nonparametric Bayesian latent factor models focuses on conjugate linear Gaussian models, for example, beta process factor analysis \citep{paisley2009nonparametric}. However, such models are not appropriate for problems where non-negativity should be imposed. To address this limitation, \cite{liang2013beta} proposed beta process NMF model by introducing a binary mask, the same as in \cite{paisley2009nonparametric}, and adopted Laplace approximation variational inference \citep{wang2013variational} for this non-conjugate model. 

However, Gaussian likelihood model was chosen for mathematical convenience; in order to perform inference, numerical optimization is required, which is computationally intensive.
Besides the computational burden, naively applying mean-field variational inference to beta process NMF model breaks the strong dependencies among the binary mask, the latent components, and the activations, and it introduces additional local optima \citep{wainwright2008graphical}. The stochastic structured mean-field (SSMF) variational inference \citep{DBLP:journals/corr/Hoffman14} was recently developed as an attempt to restore the dependencies among latent variables in the approximating variational distribution. In this paper, we utilize SSMF to address two problems: First, we develop an inference algorithm for beta process NMF models that are inherently non-negative, which, to our knowledge, has not been derived before. Second, we explore the benefit of restoring dependencies among latent variables via the SSMF framework on the task of blind source separation. %and document modeling.

\section{The model: truncated beta process non-negative matrix factorization}\label{sec:model}
In this paper, we will be working with the NMF model with Poisson likelihood, which corresponds to the widely-used generalized Kullback-Leibler divergence loss function (referred as KL-NMF). We plug the finite approximation to the beta process from \citet{paisley2009nonparametric} into the KL-NMF model:
\[
\mathbf{X} \approx \mathbf{W} (\mathbf{H} \odot \mathbf{S}) 
\]
where $\odot$ denotes the Hadamard product. Here $\mathbf{X} \in \mathbb{N}_+^{F \times T}$ represents the input data (e.g. properly scaled and quantized audio spectra), $\mathbf{W} \in \mathbb{R}_+^{F \times K}$ represents the latent components with $K$ items. $\mathbf{H} \in \mathbb{R}_+^{K \times T}$ represents the (unmasked) activations and $\mathbf{S} \in \{0, 1\}^{K \times T}$ represents the binary mask which is sparsely constructed. Concretely, the model is formulated as follows:
\begin{equation}\label{eq:model}
\begin{split}
 W_{fk} \sim \text{Gamma}(a, b); &\quad H_{kt} \sim \text{Gamma}(c, d);\\
 \pi_k \sim \textrm{Beta}(\textstyle\frac{a_0}{K}, \frac{b_0 (K-1)}{K}); &\quad S_{kt} \sim \textrm{Bernoulli}(\pi_k);\\
X_{ft} \sim \text{Poisson}&(\textstyle \sum_k W_{fk} H_{kt} S_{kt}).
\end{split}
\end{equation}
We will have a better approximation to the beta process if $K$ is set to a large value. To make inference easier, a standard trick is to introduce auxiliary random variables $\mathbf{Z}\in \mathbb{N}^{F\times T \times K}$, making use of the additive property of the Poisson random variables:
\begin{align*}
Z_{ftk} \sim &~ \text{Poisson}(W_{fk} H_{kt} S_{kt})\\
X_{ft} = &~ \textstyle\sum_k Z_{ftk}
\end{align*}
$Z_{ftk}$ can be intuitively understood as the ``contribution'' from the $k$th component for $X_{ft}$. By introducing these auxiliary random variables, when conditioning on the binary mask $\mathbf{S}$, the model enjoys the conditional conjugacy, which will be helpful when we derive the SSMF algorithm below. 

\section{Stochastic structured mean-field variational inference}

Following the stochastic structure mean-field variational inference framework, we divide the latent random variables into local: $\{\mathbf{Z}_t, \bm{s}_t\}_{t=1}^T$ and global: $\{\mathbf{W}, \mathbf{H}, \bm\pi\}$. We choose the following structured variational distribution to approximate the true posterior.
\begin{equation*}
p(\mathbf{Z}, \mathbf{W}, \mathbf{H}, \mathbf{S}, \bm\pi| \mathbf{X}) \approx q(\mathbf{Z}, \mathbf{W}, \mathbf{H}, \mathbf{S}, \bm\pi) = \Big(\textstyle\prod_k q(\bm{w}_k) q(\bm{h}_k) q(\pi_k)\Big) \Big(\prod_t q(\mathbf{Z}_t, \bm{s}_{t} | \mathbf{W}, \mathbf{H}, \bm\pi) \Big)
\end{equation*}
where the variational distributions on latent components and activations are completely factorized:
\[
q(\bm{w}_k) = \textstyle\prod_f q(W_{fk}); \quad q(\bm{h}_k)= \prod_t q(H_{kt})
\]
and take the following forms:
\[
q(W_{fk}) = \text{Gamma}(\nu^W_{fk}, \rho^W_{fk});~ q(H_{kt})= \text{Gamma}(\nu^H_{kt}, \rho^H_{kt}); ~ q(\pi_k) = \text{Beta}(\alpha^\pi_k, \beta^\pi_k)
\]
Comparing with the regular mean-field where the variational distributions are completely factorized among $\mathbf{W}$, $\mathbf{H}$, and $\mathbf{S}$, here we allow the approximated joint posterior of binary mask $\bm{s}_t$ and auxiliary variables $\mathbf{Z}_t$ to depend on $\mathbf{W}$ and $\mathbf{H}$ for each $t \in \{1, \cdots, T\}$. 
The evidence lower bound (ELBO):
\begin{align*}
\mathcal{L} &\equiv \mathbb{E}_q [ \log\textstyle \frac{p(\mathbf{W}, \mathbf{H}, \bm\pi)}{q(\mathbf{W}, \mathbf{H}, \bm\pi)}] 
+ \sum_t \mathbb{E}_q[\log \frac{p(\bm{x}_t, \mathbf{Z}_t, \bm{s}_t | \mathbf{W}, \mathbf{H}, \bm\pi)}{q(\mathbf{Z}_t, \bm{s}_{t} | \mathbf{W}, \mathbf{H}, \bm\pi)}] \leq \log p(\mathbf{X})
\end{align*}
As noted in \cite{DBLP:journals/corr/Hoffman14}, the second term corresponds to the ``local ELBO'':
\[
\mathcal{L}_t \equiv \mathbb{E}_q[\log p(\bm{x}_t, \mathbf{Z}_t, \bm{s}_t | \mathbf{W}, \mathbf{H}, \bm\pi)] - \mathbb{E}_q [\log q(\mathbf{Z}_t, \bm{s}_{t} | \mathbf{W}, \mathbf{H}, \bm\pi)] \leq \log p(\bm{x}_t | \mathbf{W}, \mathbf{H}, \bm\pi)
\]
The basic idea behind SSMF is that we can first sample global parameters from the variational distribution and then optimize the local ELBO (with respect to the local parameters) using these sampled global parameters, followed by taking a (natural) gradient step on the global parameters. This local ELBO will reach the optimum if $q(\mathbf{Z}_t, \bm{s}_{t} | \mathbf{W}, \mathbf{H}, \bm\pi)$ equals the exact conditional $p(\mathbf{Z}_t, \bm{s}_{t} | \bm{x}_t, \mathbf{W}, \mathbf{H}, \bm\pi)$, which is intractable to compute. Fortunately, SSMF only requires that we get a sample from it to construct a noisy gradient. We will resort to Collapsed Gibbs sampling to sample $\bm{s}_t$ by marginalizing out $\mathbf{Z}_t$.

\subsection{Collapsed Gibbs sampler for $\bm{s}_t$} \label{sec:cgibbs}
The construction of the auxiliary variables $\mathbf{Z}_t$ makes them straight-forward to marginalize. We can then derive the proportion of $S_{kt}$ being active or not by computing the following two quantities (Define $\hat{X}_{ft}^{\neg k} = \sum_{l\neq k}W_{fl} H_{jt} S_{lt}$), 
\begin{align*}
\mathbb{P}(S_{kt} = 1 | S_{\neg k, t}, \bm{x}_t, \mathbf{W}, \mathbf{H}, \bm\pi) 
&\propto \pi_k \cdot p(\bm{x}_t | \mathbf{W}, \bm{h}_t, S_{\neg k, t}, {S_{kt}=1}) \\
%&= \pi_k \cdot \prod_f \frac{(\hat{X}_{ft}^{\neg k} + W_{fk} H_{kt})^{X_{ft}}}{X_{ft}!} \exp\{-\hat{X}_{ft}^{\neg k} - W_{fk} H_{kt}\} \equiv P_1\\
&\propto \pi_k \cdot \textstyle \prod_f ({\hat{X}_{ft}^{\neg k} + W_{fk} H_{kt}})^{X_{ft}} \exp\{ - W_{fk} H_{kt}\} \equiv P_1\\
\mathbb{P}(S_{kt} = 0 | S_{\neg k, t}, \bm{x}_t, \mathbf{W}, \mathbf{H}, \bm\pi) 
&\propto (1 - \pi_k) \cdot p(\bm{x}_t | \mathbf{W}, \bm{h}_t, S_{\neg k, t}, {S_{kt}=0}) \\
%&= (1 - \pi_k) \cdot \prod_f \frac{(\hat{X}_{ft}^{\neg k})^{X_{ft}} }{X_{ft}!} \exp\{-\hat{X}_{ft}^{\neg k} \} \equiv P_2\\
&\propto (1 - \pi_k) \cdot \textstyle \prod_f ({\hat{X}_{ft}^{\neg k} })^{X_{ft}} \equiv P_2
\end{align*}
Finally, we can sample $S_{kt} \sim \text{Bernoulli}(\frac{P_1} {P_1 + P_2})$ after the burn-in period. To recover the per-component contribution $\mathbf{Z}_t$, note that by the property of the Poisson distribution, the conditional is multinomial-distributed: $\bm{z}_{ft} | X_{ft}, \bm{w}_f, \bm{h}_t, \bm{s}_t \sim \text{Multi}(\bm{z}_{ft}; X_{ft}, \bm\phi_{ft})$ where $\phi_{ftk} \propto W_{fk} H_{kt} S_{kt}$. Thus, we can use the conditional expectation $\mathbb{E}[Z_{ftk} | X_{ft}, W_{fk}, H_{kt}, S_{kt}] = X_{ft} \phi_{ftk}$ as a proxy.

\subsection{Update global parameters $\mathbf{W}, \mathbf{H}, \bm\pi$}

By introducing the auxiliary variables $\mathbf{Z}$, the model in Equation \ref{eq:model} enjoys the conditional conjugacy when conditioning on the binary mask $\mathbf{S}$. Therefore, the full global posterior can be factorized into conjugate pairs with respect to $\mathbf{W}$, $\mathbf{H}$, and $\bm\pi$ separately. Applying SSMF\footnote{For simplicity, we actually applied an approximated version of SSMF (referred as ``SSMF-A'' in \cite{DBLP:journals/corr/Hoffman14}).} on the corresponding variational parameters, we can obtain the full SSMF variational inference algorithm as described in Algorithm \ref{alg:ssmf}. 

\begin{algorithm}[t]
\caption{SSMF-A for beta process NMF}\label{alg:ssmf}
\begin{algorithmic}%[1]
\STATE Randomly initialize variational parameters $\{\bm\nu^W, \bm\rho^W, \bm\nu^H, \bm\rho^H, \bm\alpha^\pi, \bm\beta^\pi\}$\vspace{1pt}
\FOR{$i=1, 2, \dots$}\vspace{1pt}
\STATE Sample $W_{fk}^{(i)} \sim \text{Gamma}(\nu^W_{fk}, \rho^W_{fk})$.
\STATE Sample $H_{kt}^{(i)} \sim \text{Gamma}(\nu^H_{kt}, \rho^H_{kt})$. 
\STATE Sample $\pi^{(i)}_k \sim \text{Beta}(\alpha^\pi_k, \beta^\pi_k)$.
\STATE Sample $S^{(i)}_{kt}$ using Gibbs sampler in Section \ref{sec:cgibbs} and compute $\phi^{(i)}_{ftk} = \frac{W^{(i)}_{fk} H^{(i)}_{kt} S^{(i)}_{kt}}{\sum_l W^{(i)}_{fl} H^{(i)}_{lt} S^{(i)}_{lt}}$.
\STATE Set step-size  $\eta^{(i)} = i^{-0.5}$ and update the variational parameters:
\begin{align*}
 \nu_{fk}^W&\leftarrow (1 - \eta^{(i)}) \nu_{fk}^W + \eta^{(i)} (a + \textstyle\sum_t X_{ft} \phi^{(i)}_{ftk})\\
\rho_{fk}^W&\leftarrow (1 - \eta^{(i)}) \rho_{fk}^W + \eta^{(i)}(b + \textstyle\sum_t H^{(i)}_{kt} S^{(i)}_{kt})\\
\nu_{kt}^H&\leftarrow (1 - \eta^{(i)}) \nu_{kt}^H + \eta^{(i)} (c + \textstyle\sum_f X_{ft} \phi^{(i)}_{ftk})\\
\rho_{kt}^H&\leftarrow (1 - \eta^{(i)}) \rho_{kt}^H + \eta^{(i)} (d + S^{(i)}_{kt} \textstyle\sum_f W^{(i)}_{fk} )\\
\alpha_k^\pi&\leftarrow (1 - \eta^{(i)}) \alpha_k^\pi + \eta^{(i)} (\textstyle\frac{a_0}{K} + \textstyle\sum_t S^{(i)}_{kt})\\
\beta_k^\pi&\leftarrow (1 - \eta^{(i)}) \beta_k^\pi + \eta^{(i)}(\textstyle\frac{b_0 (K-1)}{K} + T - \textstyle\sum_t S^{(i)}_{kt})
\end{align*}
\ENDFOR
\end{algorithmic}
\end{algorithm}

\section{Experimental results}

We evaluated the proposed SSMF variational inference algorithm on synthetic examples for sanity check, as well as on real data on the task of blind source separation (BSS). In the BSS task, for comparison, we also derived a Gibbs sampler, which is slower but asymptotically exact, as an upper bound on how well the inference can potentially be. The Gibbs sampler is briefly described below.

\paragraph{Gibbs sampling}

We sampled $\bm{s}_t$ and estimated $\mathbf{Z}_t$ the same way we did for SSMF, as in Section \ref{sec:cgibbs}. The same complete conditionals can be adopted for the global parameters, which leads to Algorithm \ref{alg:gibbs}. Notice the similarity between Gibbs sampler and SSMF for this particular model: if we set the step-size $\eta^{(i)}$ in SSMF to $1$, then SSMF effectively transitions into Gibbs sampler, yet we also lose the convergence guarantee on the stochastic optimization procedure.   

\begin{algorithm}[t]
\caption{Gibbs sampler for beta process NMF}\label{alg:gibbs}
\begin{algorithmic}%[1]
\STATE Randomly initialize $\mathbf{W}$, $\mathbf{H}$, $\mathbf{S}$, and $\bm\pi$. \vspace{1pt}
\FOR{$i=1, 2, \dots$}\vspace{1pt}
\STATE Sample $S_{kt}$ using Gibbs sampler in Section \ref{sec:cgibbs} and compute $\phi_{ftk} = \frac{W_{fk} H_{kt} S_{kt}}{\sum_l W_{fl} H_{lt} S_{lt}}$.
\STATE Sample $W_{fk} \sim \text{Gamma}(a + \textstyle\sum_t X_{ft} \phi_{ftk},~ b + \textstyle\sum_t H_{kt} S_{kt})$.
\STATE Sample $H_{kt} \sim \text{Gamma}(c + \textstyle\sum_f X_{ft} \phi_{ftk}, ~ d +  S_{kt} \textstyle\sum_f W_{fk})$. 
\STATE Sample $\pi_k \sim \text{Beta}(\textstyle\frac{a_0}{K} + \textstyle\sum_t S_{kt}, ~ \textstyle\frac{b_0 (K-1)}{K} + T - \textstyle\sum_t S_{kt})$.
\ENDFOR
\end{algorithmic}
\end{algorithm}

\subsection{Synthetic data}\label{sec:synth}
We randomly sampled synthetic data following the generative process: 
We first sampled the hyperparameters: $A_{fl}, B_{fl} \sim \text{Gamma}(1, 1)$, $C_{lt}, D_{lt} \sim \text{Gamma}(5, 5)$, $\pi_l \sim \text{Beta}(0.05, 0.95)$, for $f \in \{1, \cdots, 75\}$, $t \in \{1, \cdots, 1000\}$, and $l \in \{1, \cdots, 100\}$. Then we sampled the latent variables: $W_{fl} \sim \text{Gamma}(A_{fl}, B_{fl})$, $H_{lt} \sim \text{Gamma}(C_{lt}, D_{lt})$, and $S_{lt} \sim \text{Bernoulli}(\pi_l)$. Finally the data was sampled $X_{ft} \sim \text{Poisson}(\sum_l W_{fl} H_{lt} S_{lt})$. Only 20 out of 100 $\pi_l$'s are significantly greater than 0. 

We fit the model with the hyperparameters $a = b = 0.5$, $c = d = 5$, $a_0 = b_0 = 1$, and truncation level $K= 500$. After the algorithm converged, roughly 20 out of 500 $\pi_k$'s had values significantly greater than 0, and the synthetic data was clearly recovered from the posterior mean. 

We synthesized a short clip of audio with 5 distinct piano notes and 5
distinct clarinet notes using \emph{ChucK}\footnote{\url{http://chuck.stanford.edu/}} which is
based on physical models of the instruments. At any given time,
one piano note and one clarinet note are played simultaneously at
different pitches\footnote{The clip can be listened to: \url{http://www.ee.columbia.edu/~dliang/files/demo.mp3}}. 

\begin{figure*}
  \centering
    \includegraphics[width=\textwidth]{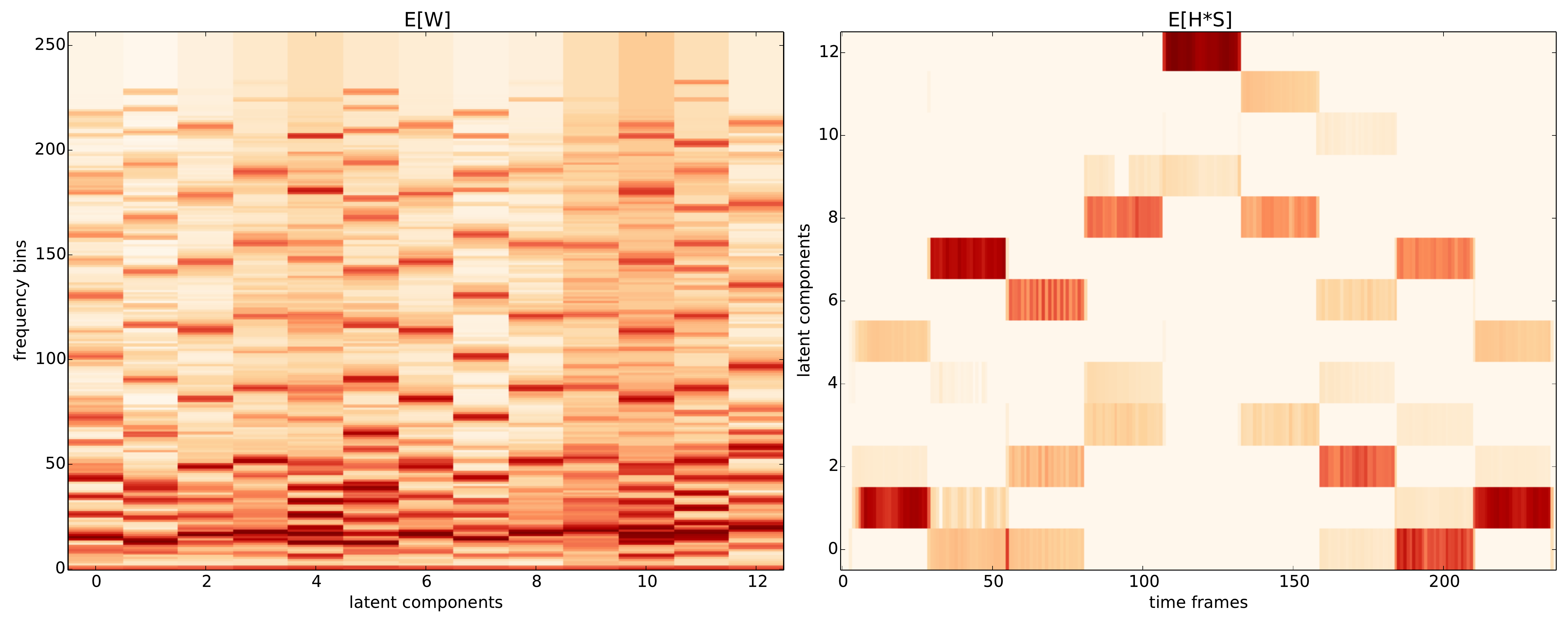}
      \caption{The NMF decomposition results on synthesized audio clip: the posterior mean (using variational distribution as a proxy) for latent components $\mathbf{W}$ (left) and activations $\mathbf{H\odot S}$ (right).}
      \label{fig:real}
\end{figure*}

The audio clip was resampled to 22.05 kHz and we computed Fast Fourier Transform (FFT) of 512 points (23.2ms) with 50\% overlap, which yielded a matrix of 257 by 238. We fit the model using the same hyperparameter setting as used above. The NMF decomposition results are illustrated in Figure \ref{fig:real}. Here we show the posterior mean (using variational distribution as a proxy) for the latent components $\mathbf{W}$ (left) and the activations $\mathbf{H\odot S}$ (right). Only the components with $\pi_k$ significantly greater than 0 are included. As we can see, the learned latent components have clear harmonic structure and capture the notes which are activated at different time. This is also implicitly reflected by the clear patterns from the activations $\mathbf{H\odot S}$ on the right. 

\subsection{Blind source separation}

We compared the performance of SSMF and Gibbs sampler on the task of audio blind source separation. We used MIREX $F_0$ estimation data, a woodwind quintet recording, which consists of bassoon, clarinet, flute, horn, and oboe. The goal is to separate individual instruments (sources) from the mixture audio signals. We resampled the audio to $22.05$ kHz and computed FFT of 1024 samples with 50\% overlap. We fit the model for both SSMF and Gibbs sampler with the hyperparameters $a = b = 0.5$, $c = d = 5$, $a_0 = b_0 = 1$, and truncation level $K= 500$. For the Gibbs sampler, we ran 200 iterations as burn-in. Given the fairly random behavior of the binary mask, we did not take multiple samples and average, instead we only took one sample for the binary mask $\mathbf{S}$. 

There is no direct information to determine how the latent components and instruments correspond, thus we used the heuristic in \cite{liang2013beta}: for each instrument, we picked the single component whose corresponding activation $\{\bm{h}_t \odot \bm{s}_t\}_{t=1}^T$ had the largest
correlation with the power envelope of the single-track instrument signal. To recover the time-domain signals, we used the standard Wiener filter.

The \texttt{bss\_eval} \citep{vincent2006performance} was used to quantitatively evaluate the separation performance. Table \ref{tab:bss} lists the average SDR (Source to Distortion Ratio), SIR (Source to Interferences Ratio), and SAR (Sources to Artifacts Ratio) across instruments
for SSMF and Gibbs sampler (higher ratios are better). As we can see, Gibbs sampler yielded better separation performance (even better than the ones reported in \cite{liang2013beta}) with running time about twice as long\footnote{The time for Gibbs sampler only includes 200 iterations as burn-in.}. The higher SIR by Gibbs sampler may be partially due to that more components were discovered (the last column), introducing less interference between components. 

On the other hand, comparably better results were obtained for SSMF than \cite{liang2013beta}, while admittedly the model assumptions are slightly different. Ideally it would be more convincing if we can also compare against the regular mean-field method. However, the binary mask $\mathbf{S}$ breaks the conditional conjugacy and we cannot directly apply the mean-field variational inference. A workaround is to use a degenerate delta function as the variational distribution for $\mathbf{S}$, as used in \cite{gopalan2014bayesian}. This will effectively estimate $\mathbf{S}$ via \emph{maximum a posteriori} (MAP), possibly with numerical optimization involved. This will be part of the future work.

The relative time difference may also come from the implementation details, but the similarity between Gibbs sampler and SSMF ensures this should not be a contributing factor.   

\begin{table}[t]
\caption{Instrument-level \texttt{bss\_eval} results with standard error in the parenthesis.  The last column represents the number of components whose corresponding $\pi_k$'s are significantly greater than 0.}
\vskip 0.15in
\begin{center}
\begin{tabular}{c | c c c | c }
\hline
  & SDR & SIR & SAR & K\\
\hline
SSMF    & 1.84 (0.92) & 6.95 (1.80) & 4.82 (0.32) & 37 \\
Gibbs &  3.58 (1.22) & 13.46 (3.69) & 5.47 (1.26) & 60  \\
\hline
\end{tabular}
\end{center}
\vskip -0.1in
\label{tab:bss}
\end{table}

%\subsection{Document modeling}

\section{Conclusion and discussion}

We present a stochastic structured mean-field variational inference algorithm for beta process KL-NMF model, which is infamously vulnerable to local optima. On synthetic examples, the model can reasonably recover the hidden structure. On a blind source separation task, SSMF performs on par with the asymptotically exact Gibbs sampler. 

There is one caveat regarding the hyperparameters. The model \emph{a priori} has two \emph{independent} channels to impose sparsity on the activation: the binary mask $\mathbf{S}$ and the unmasked activation $\mathbf{H}$. Particularly, when the prior on $\mathbf{H}$ is sparse (i.e. with $c < 1$), which produces sparse $\mathbf{H}$, $\mathbf{S}$ can always turn on a few factors (with the corresponding $\pi_k$ being close to 1) and leave the remaining factors almost completely off. This is not a desirable scenario, since we hope the sparsity pattern is captured by the binary mask. Therefore, in the experiments, we set the prior for $\mathbf{H}$ to be relatively dense with $c = d = 5$ to encourage the binary mask be sparse to ``mask out'' the dense $\mathbf{H}$. As an alternative, negative-binomial process Poisson factor model \citep{zhou2012beta} can be adopted where $\mathbf{H} \odot \mathbf{S}$ is modeled together as a random drawn from a negative-binomial distribution and efficient variational inference can thus be derived. 

\bibliographystyle{abbrvnat}
\bibliography{bpnmf_bib}

\begin{thebibliography}{10}
\providecommand{\natexlab}[1]{#1}
\providecommand{\url}[1]{\texttt{#1}}
\expandafter\ifx\csname urlstyle\endcsname\relax
  \providecommand{\doi}[1]{doi: #1}\else
  \providecommand{\doi}{doi: \begingroup \urlstyle{rm}\Url}\fi

\bibitem[Gopalan et~al.(2013)Gopalan, Hofman, and Blei]{gopalan2013scalable}
P.~Gopalan, J.~Hofman, and D.~Blei.
\newblock Scalable recommendation with {P}oisson factorization.
\newblock \emph{arXiv preprint arXiv:1311.1704}, 2013.

\bibitem[Gopalan et~al.(2014)Gopalan, Ruiz, Ranganath, and
  Blei]{gopalan2014bayesian}
P.~Gopalan, F.~J. Ruiz, R.~Ranganath, and D.~M. Blei.
\newblock Bayesian nonparametric poisson factorization for recommendation
  systems.
\newblock In \emph{Proceedings of the Seventeenth International Conference on
  Artificial Intelligence and Statistics}, pages 275--283, 2014.

\bibitem[Hoffman(2014)]{DBLP:journals/corr/Hoffman14}
M.~D. Hoffman.
\newblock Stochastic structured mean-field variational inference.
\newblock \emph{CoRR}, abs/1404.4114, 2014.
\newblock URL \url{http://arxiv.org/abs/1404.4114}.

\bibitem[Liang et~al.(2013)Liang, Hoffman, and Ellis]{liang2013beta}
D.~Liang, M.~Hoffman, and D.~P.~W. Ellis.
\newblock Beta process sparse nonnegative matrix factorization for music.
\newblock In \emph{Proceedings of the International Society for Music
  Information Retrieval Conference}, pages 375--380, 2013.

\bibitem[Paisley and Carin(2009)]{paisley2009nonparametric}
J.~Paisley and L.~Carin.
\newblock Nonparametric factor analysis with beta process priors.
\newblock In \emph{Proceedings of the 26th Annual International Conference on
  Machine Learning}, pages 777--784, 2009.

\bibitem[Smaragdis and Brown(2003)]{smaragdis2003non}
P.~Smaragdis and J.~C. Brown.
\newblock Non-negative matrix factorization for polyphonic music transcription.
\newblock In \emph{Applications of Signal Processing to Audio and Acoustics,
  2003 IEEE Workshop on.}, pages 177--180. IEEE, 2003.

\bibitem[Vincent et~al.(2006)Vincent, Gribonval, and
  F{\'e}votte]{vincent2006performance}
E.~Vincent, R.~Gribonval, and C.~F{\'e}votte.
\newblock Performance measurement in blind audio source separation.
\newblock \emph{Audio, Speech, and Language Processing, IEEE Transactions on},
  14\penalty0 (4):\penalty0 1462--1469, 2006.

\bibitem[Wainwright and Jordan(2008)]{wainwright2008graphical}
M.~J. Wainwright and M.~I. Jordan.
\newblock Graphical models, exponential families, and variational inference.
\newblock \emph{Foundations and Trends{\textregistered} in Machine Learning},
  1\penalty0 (1-2):\penalty0 1--305, 2008.

\bibitem[Wang and Blei(2013)]{wang2013variational}
C.~Wang and D.~M. Blei.
\newblock Variational inference in nonconjugate models.
\newblock \emph{Journal of Machine Learning Research}, 14:\penalty0 899--925,
  2013.

\bibitem[Zhou et~al.(2012)Zhou, Hannah, Dunson, and Carin]{zhou2012beta}
M.~Zhou, L.~Hannah, D.~B. Dunson, and L.~Carin.
\newblock Beta-negative binomial process and {P}oisson factor analysis.
\newblock In \emph{International Conference on Artificial Intelligence and
  Statistics}, pages 1462--1471, 2012.

\end{thebibliography}

\end{document}